\documentclass{tADR2e}
\usepackage{color}
\makeatletter
\newcommand\footnoteref[1]{\protected@xdef\@thefnmark{\ref{#1}}\@footnotemark}
\makeatother

\usepackage{epsfig} % for postscript graphics files
\usepackage{amsmath} % assumes amsmath package installed
\usepackage{amssymb}  % assumes amsmath package installed
\usepackage{algorithm}
\usepackage{algorithmic}

\begin{document}
\jvol{00} \jnum{00} \jyear{2015} \jmonth{January}
\articletype{SURVEY PAPER}
\title{Symbol Emergence in Robotics: A Survey}

\author{Tadahiro Taniguchi$^{a}$$^{\ast}$\thanks{$^\ast$Corresponding author. Email: taniguchi@ci.ritsumei.ac.jp \vspace{6pt}}, Takayuki Nagai$^{b}$, Tomoaki Nakamura$^{b}$,\\ Naoto Iwahashi$^{c}$, Tetsuya~Ogata$^{d}$, and Hideki~Asoh$^{e}$\\\vspace{6pt}%
$^{a}${\em{College of Information Science and Engineering,\\ Ritsumeikan University,\\ 1-1-1 Noji Higashi, Kusatsu, Shiga 525-8577, Japan}}\\%
$^{b}${\em{Department of Mechanical Engineering and Intelligent Systems,\\ The University of Electro-Communications,\\ 1-5-1 Chofugaoka, Chofu-shi, Tokyo 182-8585, Japan}}\\%
$^{c}${\em{Faculty of Computer Science and Systems Engineering,\\ Okayama
Prefectural University,\\ 111 Kubogi, Soja-shi, Okayama 719-1197, Japan}}\\%
$^{d}${\em{Department of Intermedia Art and Science,\\ School of Fundamental Science and Engineering, Waseda University,\\ 3-4-1 Ohkubo, Shinjuku, Tokyo 169-8555, Japan}}\\%
$^{e}${\em{Artificial Intelligence Research Center,\\ National Institute
of Advanced Industrial Science and Technology (AIST),\\ AIST Tsukuba
Central 1, 1-1-1 Umezono, Tsukuba, Ibaraki 305-8560, Japan}}\\%
\vspace{6pt}\received{v1.0 released ****} }

\maketitle

\begin{abstract}
Humans can learn the use of language through physical interaction with their environment and semiotic communication with other people. It is very important to obtain a computational understanding of how humans can form a symbol system and obtain semiotic skills through their autonomous mental development. Recently, many studies have been conducted on the construction of robotic systems and machine-learning methods that can learn the use of language through embodied multimodal interaction with their environment and other systems. Understanding human social interactions and developing a robot that can smoothly communicate with human users in the long term, requires an understanding of the dynamics of symbol systems and is crucially important. The embodied cognition and social interaction of participants gradually change a symbol system in a constructive manner. In this paper, we introduce a field of research called {\em symbol emergence in robotics (SER)}. SER is a constructive approach towards an emergent symbol system. The emergent symbol system is socially self-organized through both semiotic communications and physical interactions with autonomous cognitive developmental agents, i.e., humans and developmental robots.
Specifically, we describe some state-of-art research topics concerning SER, e.g., multimodal categorization, word discovery, and a double articulation analysis, that enable a robot to obtain words and their embodied meanings from raw sensory--motor information, including visual information, haptic information, auditory information, and acoustic speech signals, in a totally unsupervised manner. Finally, we suggest future directions of research in SER.
\end{abstract}
\def\Style{``jsaiac.sty''}
\def\BibTeX{{\rm B\kern-.05em{\sc i\kern-.025em b}\kern-.08em%
 T\kern-.1667em\lower.7ex\hbox{E}\kern-.125emX}}
\def\JBibTeX{\leavevmode\lower .6ex\hbox{J}\kern-0.15em\BibTeX}
\def\LaTeXe{\LaTeX\kern.15em2$_{\textstyle\varepsilon}}

\section{Introduction}\label{sec1}
\begin{figure}[t]
  \centering
  \includegraphics[width = 1.0\linewidth, clip]{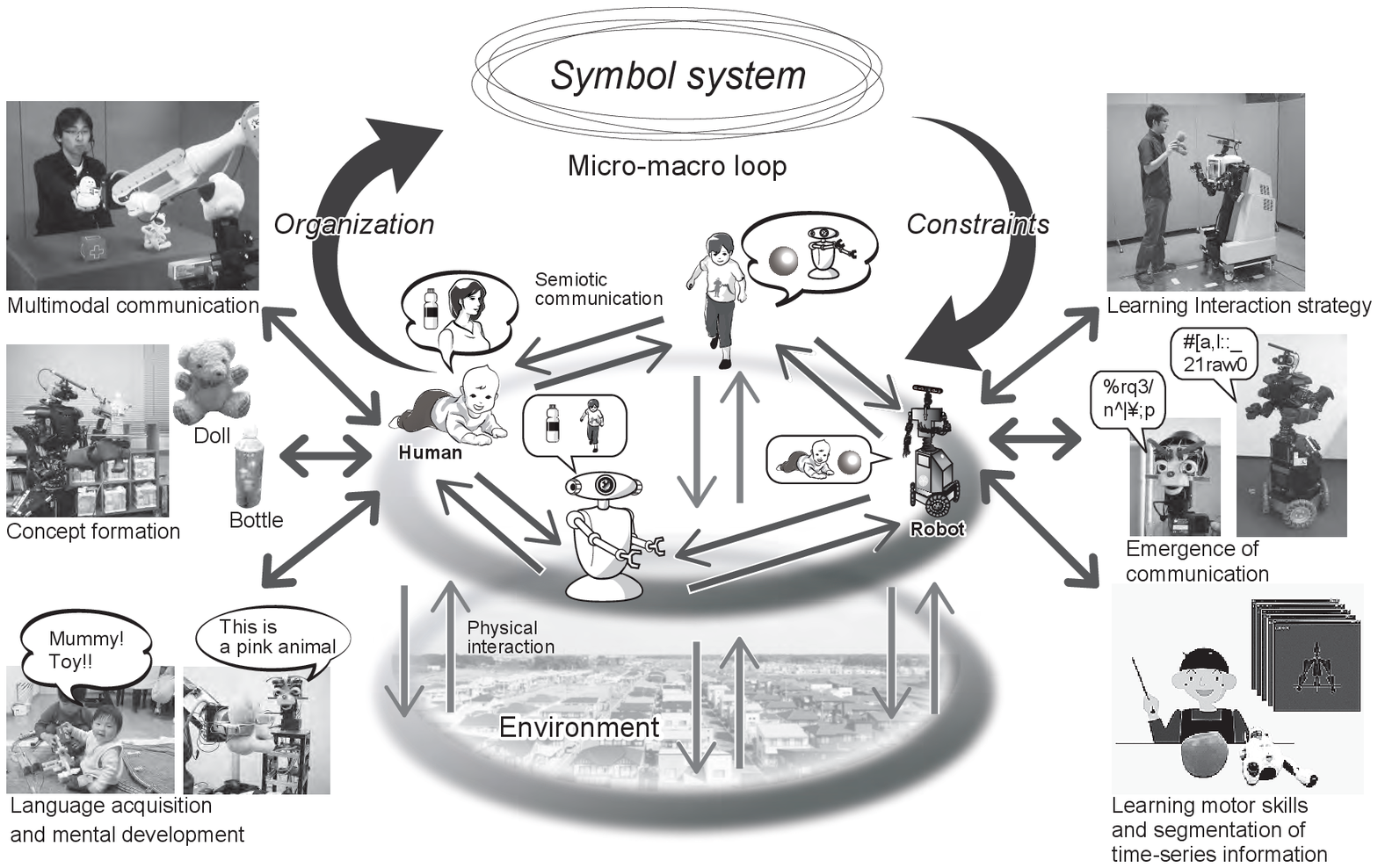}
  \caption{Emergent symbol system and research topics in symbol emergence in robotics.}
  \label{fig:ses}
\end{figure}

The development of an intelligent robot with which people would embrace a long-term interaction is one of the major challenges in the research field of robotics.
Despite the rapid and remarkable progress in robotics, natural language processing, human--robot interaction, and related artificial intelligence technologies, we have not been able to develop such an autonomous robot. Even if an entertainment robot had sufficient capabilities for speech recognition, speech synthesis, and natural language processing, a user would find it unexciting if the robot behaved deterministically on the basis of finite hand-coded rules. To overcome this problem, a robot requires the ability of open-ended development through its physical interactions and semiotic communications. Moreover, such a robot must use a language system to communicate and collaborate with humans. To achieve actual long-term communication and collaboration between a human and a robot, the robot has to understand the meanings of utterances, estimate a speaker's intention, learn new vocabularies, and promote a mutual understanding with people in the real world. To construct a mutual understanding between a human and a robot, both of them must be able to infer ``what does he/she intend by the utterance?'', ``what does the word he/she uttered represent?'', and ``what should I say to make him/her understand what I want him/her to do?'' by referring to many objects, events, contexts, situations, habits, and history. When we consider human--human communication and collaboration in the real world, it is easily understood that utterances, e.g., words, phrases, and sentences do not have one-to-one relationships with real-word phenomena. The modeling of such human semiotic communication for use in engineering applications is clearly important for developing a robot that can communicate and collaborate with others naturally like humans.

Generally, the language system is considered as a representative of the symbol systems in semiotics . 
A symbol system is an important philosophical and technical keyword, not only in semiotics, artificial intelligence, and cognitive science, but also in robotics. However, the adaptability and emergent properties of symbol systems have been underestimated and even ignored in the long history of research on robotic intelligence.

In contrast, studies that treat the adaptability of robots' internal representations and autonomous unsupervised learning processes for a language system have recently attracted attention~\cite{Cangelosi2010,Tani2014,Nakamura2009}. 
On aspect of these studies is a constructive approach to human-embodied adaptive intelligence and human symbol systems. However, the majority of current approaches to symbol systems in artificial intelligence and robotics still cannot figure out the dynamics and emergent properties of human symbol systems, i.e., the symbol systems retained in our human society. 
A philosophical theory about the dynamics of human symbol systems should be established, and a more sophisticated understanding about human symbol systems should be obtained, in order to facilitate such studies and develop intelligent robots that enable long-term communication and collaboration with humans.

Based on this notion, the research field called ``symbol emergence in robotics'' (SER) has gradually emerged over the past decade, especially in Japan. We held the first organized session concerning SER in a domestic conference in Japan in 2011.
SER is based on the concept of an ``emergent symbol system,'' which is introduced in Section~\ref{sec3}.  The emergent symbol system presumes that a human symbol system has emergent properties, and is self-organized through physical and semiotic interactions between cognitive agents, i.e., people and robots.

 Figure~\ref{fig:ses} presents an abstract figure illustrating the emergent symbol system and research topics in SER.
We assume that robots should be developed that semantically communicate with people and collaboratively interact with people in their environment, in order that they can be assimilated into the emergent symbol system.
At the center of Figure~\ref{fig:ses}, the dynamics of an emergent symbol system are schematically described. The background and concepts of emergent symbol systems are detailed in Section~\ref{sec2} and Section~\ref{sec3}.
 
 SER incorporates many research topics, such as multimodal communication, concept formation, language acquisition and mental development, learning interaction strategy, emergence of communication, and learning of motor skills and the segmentation of time-series information, as illustrated in Figure~\ref{fig:ses}.
In SER, robots are required to learn almost everything from their sensory-motor information flow, in a bottom-up manner. A top-down design of the intelligence of the robots would deprive them of the adaptability required for an emergent symbol system. In particular, concept formation that is grounded on a robot's multimodal sensory-motor information and autonomous language acquisition from row speech signals are both fundamental topics in SER. SER aims to build computational models that can describe the overall dynamics and development of language acquisition and semiotic communication, on the basis of a robot's and a child's self-enclosed sensory-motor experiences. This would enable the establishment of a new theory of embodied semantics. 

The remainder of this paper is organized as follows.
Section~\ref{sec2} briefly reviews the history of ``symbol systems'' in artificial intelligence, cognitive science, and robotics. This forms the background of emergent symbol systems, which is a philosophical prerequisite of SER. 
Section~\ref{sec3} describes the concept of emergent symbol systems.
In the subsequent sections, we provide a survey of works related to SER. In particular, we review previous studies on multimodal categorization (Section \ref{sec4}), word discovery (Section \ref{sec5}), and double articulation analysis (Section \ref{sec6}), which are important components of language acquisition, which is a fundamental challenge in SER.
In Section~\ref{sec7}, we describe further topics related to SER.
Section~\ref{sec8} concludes this paper.

\section{Background}\label{sec2}
\subsection{Physical symbol systems and robotics}
In the research field of robotics, the term ``symbol'' can be used in a variety of contexts, e.g., human--robot interaction, planning, reasoning, and communication.
Historically, the physical symbol system hypothesis was proposed by Newell and Simon~\cite{Newel1976,Newell1980}. This formed the starting point for a discussion about symbol systems in artificial intelligence and related fields. But, it was a problematic starting point. The philosophy is clearly inspired by early successes in computer science and programming languages.
Many related works that followed the physical symbol system hypothesis and/or its way of thinking have placed an emphasis on the manipulation of symbols in research on artificial intelligence. This way of thinking was inherited from the tradition of ``symbolic logic.'' 

In predicate logic, which is a representative of symbolic logic, predicates and variables that represent real-world phenomena are given as discrete representations in a top-down manner~\cite{Russell2009}. 
The fundamental assumption is that our world can be distinguished and segmented into a discrete ``symbol'' system, and that the system is deterministic and static.
In other words, predicate logic can describe the world so far as such assumptions are satisfied. This represents a type of ``approximation.'' Almost all symbolic logic essentially shares the same assumptions. Physical symbol system hypothesis, proposed by Newel, is no exception~\cite{Newel1976,Newell1980}. 

This convention has implicit effects on studies in robotics. 
In current robotics research, a ``symbol'' tends to be regarded as a ``discrete'' entity, having a ``one-to-one'' relationship with a word. A symbol is regarded as a manipulatable element in the mind, i.e., a robot's memory system. For example, in \cite{Inamura2004} the authors call a type of trajectory of a humanoid's entire bodily motion, modeled by a left-to-right hidden Markov model, a ``proto-symbol.''
Such notions as ``a symbol is a discrete component of a memory system in a robot,'' ``a symbol is an internal representation in a robot,'' and ``a symbol system is a set or a network of such components,'' have spread widely through the artificial intelligence and robotics communities. 

Figuratively speaking, symbolic logic adopts the assumptions of equilibrium and determinism for modeling an actual human symbol system. It is assumed that the human symbol system is the same for everyone, and does not change over time.
This approximation has been valid for solving many problems, in the same way that linear control theory has solved many problems, even though most real-world systems have nonlinear and stochastic properties, or that the theory of thermodynamics of equilibrium systems has provided fruitful results for engineering purposes. 

However, these assumptions are crucially problematic, and have misled many researchers over the past four decades. This has resulted in people misunderstanding the human symbol system. The blind acceptance of the approximation has meant that people did not consider the following important characteristics of the human symbol system.
\begin{description}
\item[C1]Grounded:~  A symbol does not have any meaning without being grounded or interpreted.
\item[C2]Dynamic:~ There does not exist an objectively true symbol system that can be determined in top-down manner in our human society.
\item[C3]Social:~ An individual representation system and the socially shared symbol system are not same.
\end{description}
The characteristics of symbols have been widely accepted in semiotics and in a broader context of humanities research~\cite{Chandler2002}.

\subsection{Physical grounding hypothesis}
 There have been many criticisms of the physical symbol system hypothesis, and approaches to intelligent systems based on this hypothesis, in the field of artificial intelligence. 
Brooks representatively criticized and insisted that sensory-motor coupling with the environment is primarily important for robots to achieve everyday tasks in our daily environment~\cite{Brooks1990,Brooks1991}.
His famous paper ``intelligence without representation'' provided a clear objection to the physical symbol system hypothesis. He proposed physical grounding hypothesis, of which the key observation is that ``the world is its own best model.''
 He developed many robots based on a subsumption architecture, which is a reactive and decentralized robotic architecture. This behavior-based robotics places an emphasis on the primal sensory--motor interaction between a robot's embodied system and its environment, and the emergence of behavior through interactions. Breazeal et al. even developed a ``social'' interactive robot using the subsumption architecture~\cite{Breazeal2003,Breazeal2004}. However, the subsumption architecture is still a framework for ``designing'' a robot that behaves naturally in our daily environment. It is difficult for such an approach to build an autonomous system that gradually reaches an intelligent state such that it can communicate with people using a language system, i.e., a human symbol system.

The field of embodied cognitive science is closely related to Brooks's approach~\cite{Pfeife2001}. This approach is also related to the research field of artificial life and complex systems.
Metaphorically, Brooks's approach is to develop an insect-like artifact. In contrast, the traditional approach to artificial intelligence can be thought of as an attempt to develop an obstinate mathematician who cannot behave appropriately in a real-world environment.

\subsection{Symbol grounding problem}
The other famous criticism in relation to the symbol system, was expressed by Harnad. He proposed the symbol grounding problem (SGP), which is one of the most famous problems in artificial intelligence~\cite{Harnad1990}. 
The SGP focused on the relationship between a designed symbol system and real-world phenomena. The importance of the SGP has been widely recognized over the past two and a half decades.
The design of a ``symbol system'' and application of it to an autonomous agent in a top-down manner inevitably leads to the SGP. 

Advocates of physical symbol systems have insisted that the meaning of a symbol is syntactically determined in relation with other symbols. However, such a ``relationship'' cannot reach a conclusion on what anything means. A relationship between two signifiers can never provide the relationship between a signifier and a signified object. Harnad compared this phenomenon to a ``merry-go-round''. To obtain any meaning, a word has to be grounded via sensory--motor information, or borrow meanings from other words using syntactic rules. Cangelosi et al. called these processes ``sensorimotor toil'' and ``symbolic theft'', respectively~\cite{Cangelosi2000}. 

In cognitive science, physical symbol systems have also been criticized. Barsalou proposed the concept of ``perceptual symbol systems'', to place an emphasis on perceptual experiences for theories of knowledge~\cite{Barsalou1999}. He called the static symbolic system an ``amodal symbol system,'' and pointed out its drawbacks. Although the notion of a perceptual symbol system was not completely new, Barsalou mentioned that a perceptual theory of cognition may lead to a competitive, and perhaps superior, theory. The physical symbol system clearly relates to the SGP. 

Many interdisciplinary studies have aimed to solve the SGP~\cite{Cangelosi2006,Araki2012,Tani2014,Sinapov2014}. Recently, Tellex et al. presented an approach to the SPG using probabilistic graphical models~\cite{Tellex2011}.
From a philosophical viewpoint, Taddeo et al. proposed the zero semantical commitment condition, which must be satisfied by any hypothesis seeking to solve the SGP~\cite{Taddeo2005}. 

However, despite the long history of the SGP, a clear solution has not been found. One of the reasons for this is that the SGP itself is naively defined. The SGP is based on the physical symbol system. Therefore, the SGP itself was misled by the physical symbol system hypothesis. That is, the SGP is an ill-posed problem. The SGP mainly considers {\bf C1}, almost completely ignoring {\bf C2} and {\bf C3}.
Steels pointed out the problem in an ambitious and impressive paper titled ``The symbol grounding problem has been solved. So what's next?''~\cite{Steels2008}. He described it as follows:
\begin{quote}
I propose to make a distinction between
{\em c-symbols}, the symbols of computer science, and {\em m-symbols}, the meaning-oriented symbols in the tradition of the arts, humanities, and social and cognitive sciences.
\end{quote}
This distinction is crucially important for the construction of theories concerning long-term human--robot interactions.
The SGP starts with c-symbols, and attempts to make them grounded. 
To develop an intelligent robot that people would embrace long-term interactions with, we should clearly start from m-symbols, because from a human viewpoint, a human--robot interaction is composed of m-symbols. In our paper, we call a system of c-symbols and m-symbols as an internal representation system and a human symbol system, respectively, according to the conventions of robotics and semiotics. 
For example, Weng provided a critical survey on symbolic models and emergent models in artificial intelligence~\cite{Weng2012a}. Those symbol models are about internal representation systems in our terminology.

In contrast, the emergent symbol system we introduce in the next section considers both types of symbol, in an integrative manner.

\subsection{Developmental robotics}
The epigenetic and/or developmental viewpoint is crucially important in creating artificial intelligent systems that can adapt to a dynamic real--world environment. The field of developmental robotics has emerged gradually over the past two decades~\cite{Cangelosi2015}. Cangelosi et al. described developmental robotics as follows~\cite{Cangelosi2015}:
\begin{quote}
Developmental robotics is an approach to the autonomous design of behavioral and cognitive capabilities in artificial agents (robots) that takes direct inspiration from the developmental principles and mechanisms observed in the natural cognitive systems of children.
\end{quote}
The field is also referred to as ``epigenetic robotics''~\cite{Cangelosi2006}.
Asada et al. used the term ``cognitive developmental robotics''~\cite{Asada2009}.  Asada et al. stated that cognitive developmental robotics places more emphasis on human/humanoid cognitive development than on related approaches.
Our research field, SER, philosophically inherits many concepts and fundamental assumptions from the field of (cognitive) developmental robotics. In a manner of speaking, SER is a (crucially important) branch of developmental robotics. 

Developmental robotics places an emphasis on an autonomous agent's embodied interaction with the environment and the adaptive organization of the cognitive system, including cognitive capabilities relating to language and other symbol systems~\cite{Cangelosi2015}. However, the scope of developmental robotics is tremendously large, because it involves almost all of human intelligence and its diachronic changes. Moreover, developmental robotics attaches importance to interdisciplinary communication between robotics and developmental psychology. Many efforts have been made to construct a fruitful interdisciplinary academic field. 

However, these characteristics of developmental robotics have distracted its attention from a computational and constructive understanding of dynamic human symbol systems and the development of robots that achieve the overall dynamics and development of language acquisition and semiotic communication. We believe that these are central topics in robotics research for achieving long-term human--robot communication and collaboration. This is our motivation for introducing the field of SER.

\subsection{Symbol emergence in robotics}
The approach in SER places more emphasis on the computational understanding of emergent symbol systems.  In addition to cognitive development, SER attempts to cover semiotic phenomena. The field of SER is an interdisciplinary field, which is not only related to robotics, artificial intelligence, development psychology, and cognitive science, but also to semiotics and linguistics as well.

To describe the diachronic changes in internal representation systems and human symbol systems that are caused by embodied interaction and social communication, we require mathematical models, such as generative models, neural networks and related statistical models, and robotic models such as humanoids and mobile robots, for a productive discussion and development of the integrative theory. In cognitive science, the generative probabilistic model has recently been widely used to represent the human cognitive system~\cite{Tenenbaum2011}. In addition to such computational models, SER places an emphasis on embodied cognition. Therefore, researchers in the field of SER use robotic models to connect computational models to the real physical world. This involves the use of state-of-art machine-learning technology, including Bayesian nonparametrics and deep neural networks, to model the diachronic changes in the cognitive systems and symbol systems of human/robots. 

%The main difference between SER and earlier related approaches is that SER assumes that the human symbol system is a socially self-organized system with emergent properties. We believe that assumption can be fruitful for the future development of intelligent robots. The next section introduces emergent symbol systems.

\section{Emergent Symbol Systems}\label{sec3}
The center of Figure~\ref{fig:ses} shows the schematic figure of an emergent symbol system that was originally introduced in \cite{Taniguchi2010book}. SER is defined as a constructive approach towards emergent symbol systems~\cite{Taniguchi2014book}. In this section, we explain emergent symbol systems by referring to the figure. 
\subsection{Semiosis and umwelt}
We will start from a human symbol system, i.e., the m-symbols described by Steels~\cite{Steels2008}. The preexisting interdisciplinary research field that deals with human symbol systems is called semiotics.
Semiotics is concerned with everything that can be interpreted as a sign, as explained by Eco~\cite{Eco1976}. Initially, semiology was introduced by Saussure, while Peirce independently introduced semiotics~\cite{Saussure1983,Peirce}. Currently, the two fields have overlapped, and merged into the academic field called semiotics. From the viewpoint of semiotics, language is a representative of general symbol systems. 

In Peircean semiotics, a symbol is defined as a process having three elements. The definition has a high affinity for the bottom-up approach to cognitive systems. The first is the {\it sign (representamen)}, which describes the form that the sign takes, the second is the {\it object}, which is something that the sign refers to, and the third is the {\it interpretant}, which, rather than an interpreter, is the sense made of the sign.
The important point of the Peircean definition of a symbol is that the sign, e.g., words, visual signs, or pointing, is not a symbol itself. The interpretant, the third element of a symbol, mediates between the sign and the object. This degree of freedom allows us to take a variety of interpretations and dynamics of a symbol system into consideration.
In the Peircean definition, a symbol is not a static material, but a dynamic process of interpretation. Peirce calls this process ``semiosis''~\cite{Peirce}.

The definition of a symbol is still abstract, but the definition clearly satisfies {\bf C1}, {\bf C2}, and {\bf C3} from Section~\ref{sec2}.  The utterances of others are always interpreted on the basis of semiosis in {\bf semiotic communication}.

Peircean semiotics places a thorough emphasis on the subjective viewpoint. 
Uexk\"{u}ll, who established biosemiotics, proposed the famous notion of {\it umwelt}. The umwelt represents an animal's subjective world, which emerges on the basis of the animal's sensory--motor system~\cite{Uexkull}. Brooks also cited Uexk\"{u}ll, when he attacked the physical symbol system in his famous paper~\cite{Brooks1990}.
We should start take the umwelts of robots and humans as a starting point. 
Their internal representation systems are initially formed through {\bf physical interactions} with their {\bf environment}, using their sensory--motor systems. 

In this sense, {\em semiosis} is the key that connects Brooks' physical grounding hypothesis, which eliminated internal representation systems, to semiotic communication, which is required for long-term human--robot communication.

\subsection{Arbitrariness and perspective of structurists}
In contrast to Peirce, Saussure emphasized the synchronic structure of language. 
The defining notion of Saussurean semiotics is the {\em arbitrariness} of the sign (symbol).
This embodies {\bf C2}. The relationships between signs, such as labels and words, and categories are arbitrary, and the categories and segments of phenomena are also arbitrary. The arbitrarily determined categories, segments, and lexicons are retained in a language system to which many people, who speak the language, belong. Structuralists, the successors of Saussure, place an emphasis on arbitrariness. When we belong to a language system, i.e., a human symbol system, our cognition, interpretation, utterances, and even behaviors are affected by the symbol system. For example, we comprehend objects so as to classify them into preexisting categories that our language system retains. The symbol system provides {\bf constraints} on our semiotic communication and physical interaction. 

According to structuralists, ``things'' do not exist independently of the symbol system that we use; reality is the creation of the media that seems to simply represent it.
The structuralist perspective tends to reverse the precedence of language and cognition. They stress that our language, which incorporates arbitrariness, determines the order of the world~\cite{Sturrock1986}. 
This places top-down constraints in an emergent symbol system.

\subsection{Emergent symbol systems}
This structurist exaggeration of unilateral determination is not accurate. Human symbol systems can be {\bf organized} in a bottom manner. Genetic epistemology was proposed by Piaget~\cite{Piaget1971}, who is often called the father of cognitive development research. In genetic epistemology, the subjective world of humans is considered to be gradually ``constructed'' through interactions with their environment. Piaget introduced a schema system, which is a self-organized cognitive system that emerges through sensory--motor interaction, and is believed to be the basis of the language system~\cite{Flavell}.

An internal representation system is not a static system, but rather a dynamic system that is self-organized through physical interaction based on the sensory--motor system. Furthermore, a human symbol system is organized through semiotic communication on the basis of individual internal representation systems in a bottom-up manner (see {\em organization} in Figure~\ref{fig:ses}).
In semiotics, symbol systems are seldom treated as the static, closed, and stable systems that are inherited from preceding generations, but instead are regarded as constantly changing~\cite{Chandler2002}.
The bottom-up organization and the top-down constraints of the symbol system introduce an emergent property 
to the overall system.

Once a symbol system is generated in a society, people who use the symbol system must obey the rules of this system to communicate and collaborate with others. The symbol system includes phonetics, lexicons, syntax, and pragmatics as its constituents. If an agent belonging to the society does not follow the rules, i.e., the symbol system, then the agent cannot communicate its idea to others or collaborate with others. This means that the agent cannot make use of the powerful symbolic system for their further survival. 

Such a bilateral relationship between an emerged symbol system at the macro level and a physical system consisting of communicating and collaborating agents at the micro level forms a {\bf micro-macro loop}. Micro-macro loops are found in many complex systems, especially in living systems. This tells us that the entire system is an {\em emergent system}, i.e., a complex system having emergent properties.
Polanyi, who introduced the notion of emergence, described it as follows~\cite{Polanyi1966}:
\begin{quote}
If each higher level is to control the boundary conditions left open by the operations of the next lower level, this implies that these boundary conditions are in fact left open by the operations going on at the lower level.
\end{quote}
\begin{quote}
This stratification offered a framework for defining {\it emergence} as the action that produces the next higher level, first from the inanimate to the living, and then from each biotic level to the one above it.
\end{quote}

The discussion above provides us with a novel concept, called an {\it emergent symbol system} (Figure~\ref{fig:ses}). This emergent symbol system is also closely related to an autopoietic system~\cite{Maturana}.

To develop a robot that naturally communicates and collaborate with humans in a long-term manner, we have to create a robotic system that behaves as an adaptive element of the emergent symbol system.
To become an element of the emergent symbol system, a robot has to have a capability to form an internal representation system, acquire a language system that humans use, modulate its pragmatics, communicate with people in various contexts, and autonomously collaborate with people in a physical environment. These are the same tasks that human children are required to complete during their developmental period. 
To achieve a long-term development and adaptation to an emergent symbol system, a robot should learn the symbol system in an unsupervised manner and perform communication and collaboration autonomously.

In SER, which is a constructive approach to emergent symbol systems, our initial aim should be to develop an autonomous robot that can automatically acquire language like a human child, and gradually learn to communicate and collaborate with humans in our daily environment, in a bottom-up manner.

\section{Multimodal Categorization}\label{sec4}
\begin{figure}[t]
  \centering
  \includegraphics[width = 0.9\linewidth]{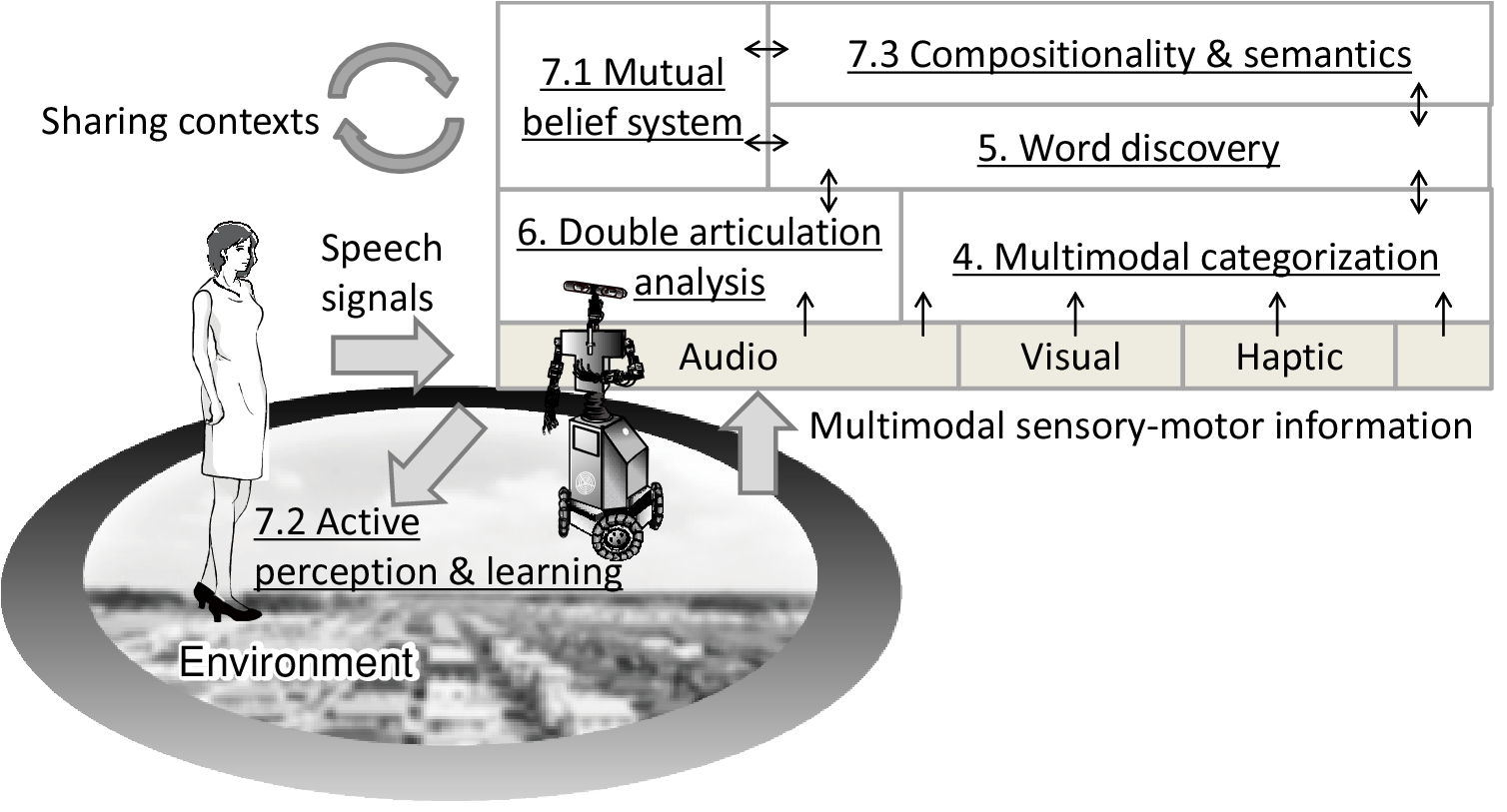}
  \caption{Mutual relationships of components corresponding to research topics introduced between Sections~\ref{sec4} and \ref{sec7}. Each arrow indicates the dependency of each component and information flow among the components.}
  \label{fig:components}
\end{figure}

From this section to Section~\ref{sec7}, we present a brief survey of practical studies related to SER.
Figure~\ref{fig:components} illustrates the relationships between the topics described in following sections. These are the basic components required to develop a robot that can learn language autonomously and become an element of an emergent symbol system.

Embodied automatic language acquisition is one of the central issues in SER. 
The development of an embodied cognitive system that can automatically acquire language is different from the development of conventional natural language processing and automatic speech recognition systems. In language acquisition, a robot must learn language in a basically unsupervised manner from its embodied sensory--motor information. Owing to the remarkable progress in machine-learning and robotic technologies, the range of realized automatic language acquisition methods is currently growing.

\subsection{Object category formation}
Before obtaining language, human children are considered to obtain object categories gradually through daily interactions with objects. Piaget insisted that the schema system self-organizes through the sensory--motor period, and that the system becomes the prerequisite for language~\cite{Flavell}. An embodied multimodal sensory--motor experience must be a primal root for human category formation.

A category formation problem is different from a pattern recognition problem. In a pattern recognition problem, truth labels for the recognition results are provided in a supervised manner. A vast number of studies have been carried out regarding the development of an accurate pattern recognizer. Recently, deep learning methods have yielded excellent results~\cite{Krizhevsky2012,Dahl2012}.
In contrast, category formation in a robot's umwelt must be autonomously performed in an unsupervised manner.
From the viewpoint of machine-learning, object categorization is regarded as a clustering problem, involving a type of unsupervised machine-learning tasks~\cite{Bishop2010}.

Historically, many studies have emphasized visual information in category formation by computational systems. However, the formation of object categories based solely on visual information is insufficient, because our human symbol system is organized on the basis of our multimodal sensory--motor experiences.
The integration of multimodal information through category formation is important for a robot to predict future sensor information. By forming an object category on the basis of visual, auditory, and haptic information, a robot can infer auditory and haptic information from the recognized category, e.g., a bottle and a cymbal, from its visual information.

\subsection{Computational models for multimodal object categorization}
Recently, various computational models for multimodal object categorization have been proposed~\cite{Celikkanat,Sinapov,Natale,Araki2012,Ando,Nakamura2007,Nakamura2009,Nakamura2011,MHDP,Nakamura2014,Griffith2012,Iwahashi2010,Roy2002a,Sinapov2014}. For example, Sinapov et al. proposed a graph-based multimodal categorization method, which allows a robot to recognize new objects on the basis of similarities to a set of familiar objects~\cite{Sinapov}. They also made a robot perform ten different behaviors; obtain visual, auditory, and haptic information; and explore 100 different objects, classifying them into 20 object categories ~\cite{Sinapov2014}.
However, their multimodal categorization is performed in a supervised manner. Celikkanat et al. modeled the context in terms of a set of concepts, allowing many-to-many relationships between objects and contexts using latent Dirichlet allocation (LDA), inspired by the notion of situated concepts introduced by Yeh and Barsalou~\cite{Wenchi2006,Celikkanat}.
Mangin used a nonnegative matrix factorization algorithm to learn a dictionary of components from multimodal time series data~\cite{Mangin2013}.
Natale et al. have demonstrated that a robot can recognize objects with the help of a self-organizing map (SOM), using proprioceptive data extracted from the robot's hand as it grasps an object~\cite{Natale}.
Lallee et al. proposed multi-modal convergence maps on the basis of SOMs. This method can integrate visual, motor, and language modalities~\cite{Lallee2013}.
Invalid et al. proposed a cognitive architecture, and developed a child-like robot that can automatically learn object categories through active exploration~\cite{Ivaldi2014a}. 

\begin{figure}[t]
  \centering
  \includegraphics[width = 1.0\linewidth]{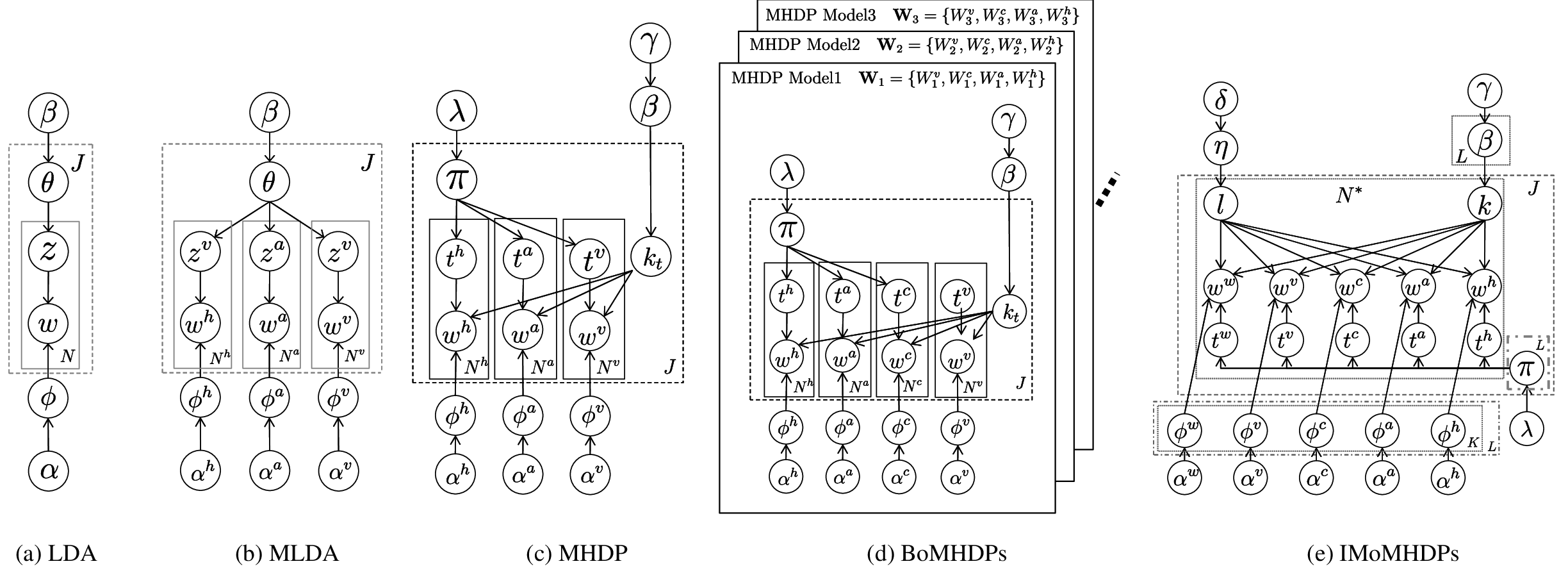}
  \caption{Graphical models for object categorization.}
  \label{fig:mlda_gm}
\end{figure}

Nakamura et al. have presented a series of studies on multimodal categorization using multimodal latent Dirichlet allocation (MLDA) and its extentions~\cite{Araki2012,Ando,Nakamura2007,Nakamura2009,Nakamura2011,MHDP,Nakamura2014,Nakamura2015}.
They have extended LDA, which was first proposed by Blei et al. for document-word clustering, to a model that can treat multimodal information~\cite{Blei2003,Griffiths2004,Nakamura2009}. Concrete illustrations of the graphical models of LDA and MLDA and its extended models are presented in Figure~\ref{fig:mlda_gm}. MLDA has several emission distributions for an object, i.e., a document in LDA. The object categories of objects in multimodal categorization in MLDA correspond to topics in document-word clustering in LDA.

\begin{figure}[t]
  \centering
  \includegraphics[width = 0.8\linewidth]{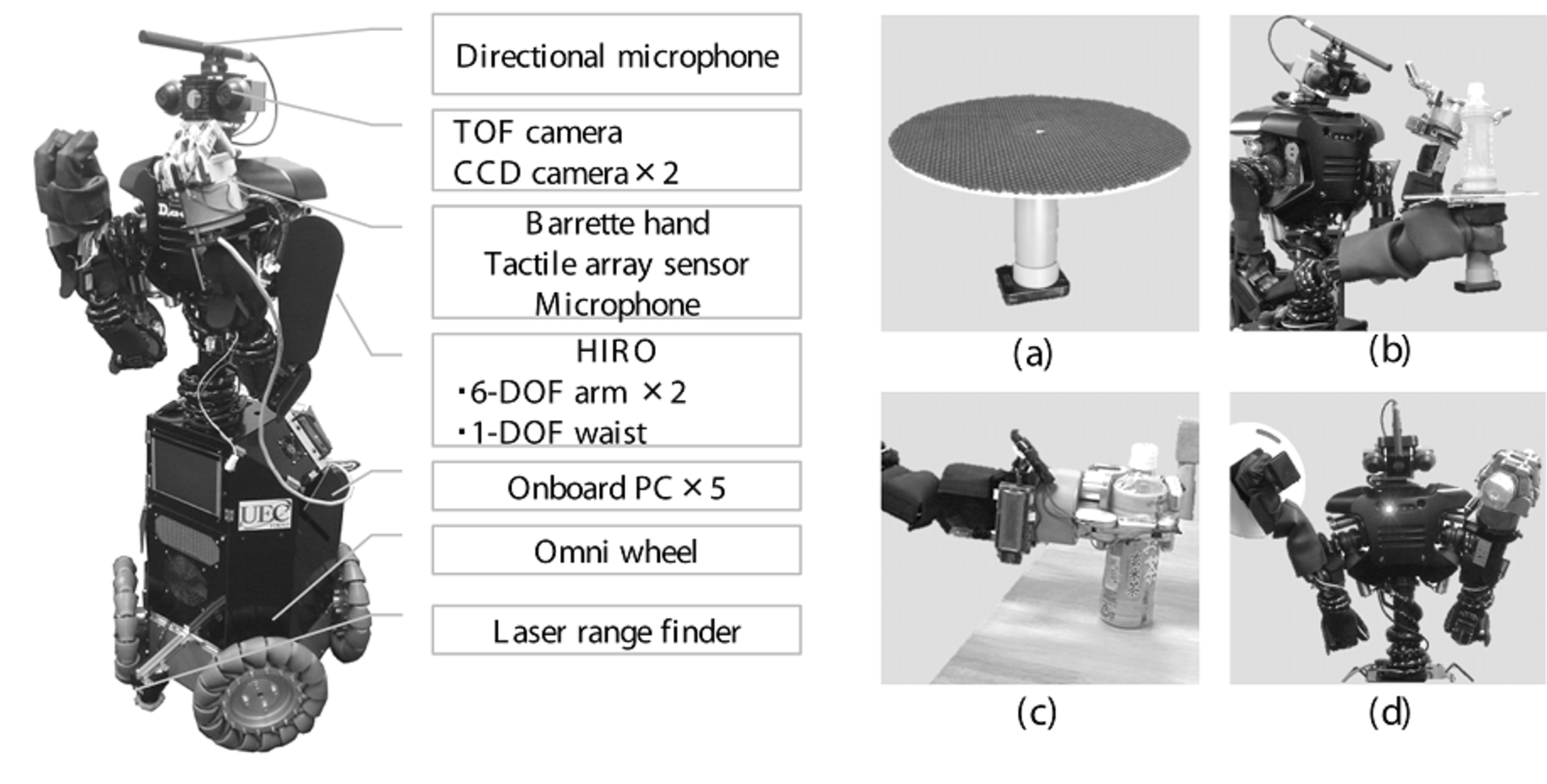}
  \caption{The robot used in multimodal categorization experiments.}
  \label{fig:daigoro}
\end{figure}

Nakamura et al. developed a robotic system that can obtain visual, audio, and haptic information by interacting with objects. An overview of the robot is presented in Figure~\ref{fig:daigoro}. The robot can grasp an object and observe it from various viewpoints. The robot has cameras, microphones, arms, and hands with pressure sensors. The robot obtains visual information by taking pictures of a target object from many directions, by rotating the object with its hand. The robot also obtains haptic information by grasping the target object several times, and audio information by shaking the target object.
Feature vectors are extracted from the observed information of each modality, and the feature vectors of each modality are transformed into bag-of-features representations using the K-means method, i.e., vector quantization. The bag-of-features representations are passed to MLDA, and the clustering procedure is performed.
Nakamura et al. demonstrated that a robot can categorize a large number of objects in a home environment into categories that are similar to human categorization results~\cite{Nakamura2009}.
Araki et al. developed online MLDA, and performed an experiment on multimodal category acquisition in a fully autonomous manner in a home environment~\cite{Araki2011}.
The result indicates that a human symbol system is not necessarily required for a cognitive system to form human-like object categories. This suggests that the human symbol system is not completely arbitrary, but has a certain rationality brought by the latent structure embedded in sensory-motor information, as admitted by Saussure~\cite{Saussure1983}.

\subsection{Estimating latent structure in multimodal categories}
Although MLDA is able to form multimodal categories, it can not adaptively estimate the number of object categories. It is unlikely that the number of categories that a cognitive system has is determined in advance. The Bayesian nonparametric approach provides a reasonable solution to the problem. Nakamura et al. extended the hierarchical Dirichlet process (HDP), which was a nonparametric Bayesian clustering method proposed by Blei et al. for document-word clustering, to multimodal HDP (MHDP), which can treat multimodal information and automatically estimate the number of categories from the observed multimodal information~\cite{HDP,Sudderth2006,MHDP}.

Bayesian nonparametrics is a branch of the Bayesian approach. In general, the number of hidden variables in the Bayesian nonparametric approach can be automatically estimated by using the infinite dimensional prior distribution, for example using Dirichlet or Beta processes, and a feasible inference procedure~\cite{HDP-HMM,HDP}. The mathematical framework is very important for constructing a computational model relating to emergent symbol systems.
Nakamura et al. demonstrated that a robot can also estimate the number of object categories to give similar results to human categorization results~\cite{MHDP}.

MLDA and MHDP can easily be extended to treat ``words.'' LDA and HDP were originally applied to document and word clustering methods~\cite{Blei2003,HDP}. By adding an observation variable for words to MLDA's graphical model, MLDA is able to cluster multimodal information and words simultaneously. As a result, a robot can estimate the label of a category in an unsupervised manner.

More complex latent structures of multimodal categories can be estimated. Ando et al. proposed hierarchical MLDA (hMLDA), by extending hierarchical LDA (hLDA) for hierarchical multimodal categorization~\cite{Blei2007,Ando}. This method enabled a robot to form a hierarchical structure of object concepts from multimodal sensory--motor information, e.g., ``plastic bottle'' is a subcategory of ``water container.'' 

Nakamura et al. proposed the Bag of MLDA (BoMLDA), Bag of the MHDP (BoMHDP), and infinite mixture of MHDPs (IMoMHDPs) methods, which can perform various types of categorization with different perspectives~\cite{Nakamura2011,Nakamura2012a,Nakamura2015}. The methods emphasize some modalities, and organize categories based on information of the modalities focused on. By adopting these methods, their robot was able to form concepts about not only ``objects'', but also ``attributes,'' e.g., soft, hard, green, red, or yellow, from multimodal information.

Compared to related methods for multimodal categorization, the MHDP-based approach is sophisticated from the viewpoint of Bayesian modeling. Its mathematical soundness and theoretical consistency help us to build new methods that are based on it, e.g., active perception~\cite{Yoshino2015}.

\subsection{Multimodal representation learning using neural networks}
Another way to integrate multimodal information involves approaches that utilize neural networks. 
Ngiam et al. applied deep networks to learn features over multiple modalities. By integrating visual, i.e., lip motions, and auditory information, they developed a robust speach classification system. They also demonstrated that the system shows McGurk effect that is an audio-visual perception phenomenon in the similar way as human~\cite{Ngiam2011}. 
Noda et al. integrated auditory, visual, and motor information using a deep neural network~\cite{Noda2013}. In their experiment, a robot could recall upper bodily motion from visual and audio information, and retrieve image information from sound and joint angle inputs. Heinrich et al. extended the multiple timescale recurrent neural network (MTRNN), and obtained multi-modal MTRNN, to integrate visual, auditory, and motor information. Recently, neural networks with a deep architecture have attracted attention. Le et al. demonstrated that large-scale unsupervised learning using a deep neural network could build high-level features automatically from image data~\cite{Le2011}. Bridging representation learning using neural networks and multimodal categorization using generative models represents important work in this field~\cite{Bengio2013}.

\section{Word Discovery}\label{sec5}
\subsection{Word discovery by human children}
In language acquisition, word discovery, i.e., word segmentation, is an important task for children. A word is an elemental pattern of a linguistic sign. A phoneme is an element that is acoustically but not semantically distinguishable. 
Discovering words from continuous speech signals is a fundamental task that children must solve to acquire language.
Unlike an automatic speech recognition system, children must learn a language model, i.e., a word inventory and transitional information about the words, and in an acoustic model, i.e., an organized memory about phonemes, this must be done from speech signals in an unsupervised manner.

What types of cue can be used by children to discover words from continuous speech signals?
Three representative cues for word segmentation are listed by Saffran et al.~\cite{Saffran1996a}.
These are {\it distributional}, {\it co-occurrence}, and {\it prosodic} cues.
Distributional cues concern the statistical relationships between neighboring speech sounds. These can be modeled as n-gram statistics to some extent, once each phoneme is recognized correctly.
Co-occurrence cues concern entities detected in the environment by children. For example, when a child hears two sentences while he/she looks at an apple, the sentences are likely to contain overlapping words, such as ``apple.''
Prosodic cues relate more to superficial acoustic information, such as stressed syllables, post-utterance pauses, and acoustically distinctive final syllables.

All of the above cues are believed to contribute to word discovery in an integrative manner. Among these, Saffran emphasized distributional cues.
She reported that word segmentation from fluent speech could be accomplished by eight-month-old infants using only distributional cues~\cite{Saffran1996}. By the age of seven months, infants are reported to use distributional cues~\cite{Thiessen2003}.

\subsection{Word discovery by robotic systems}
In SER, autonomous word discovery by robots is one of the first challenges that should be solved. In the past two decades, many types of unsupervised machine-learning methods for word discovery (segmentation) have been proposed~\cite{Brent1999,Venkataraman2001,Goldwater2008,Goldwater2009,Mochihashi2009,Johnson2009,Chen2014,Magistry2012,Sakti2011}.
Conventionally, Brent proposed model-based dynamic programming to find word boundaries in a natural-language text whose word boundaries are deleted~\cite{Brent1999}. Venkataraman proposed a statistical model to improve Brent's algorithm~\cite{Venkataraman2001}.

In contrast with such text-based approaches, Roy et al. developed a computational model and a robotic system that autonomously discovers words from a raw multimodal sensory input~\cite{Roy2002a}. The experimental results of Roy et al. demonstrated the development of a cognitive robot that can acquire a lexicon from raw sensor data without human transcription or labeling. Although imperfect, the results were encouraging. 
Their results showed that it is possible to develop cognitive models that can process raw sensor data and acquire a lexicon, without the need for human transcription or labeling.

Contemporaneously, 
Iwahashi et al. independently proposed a sophisticated probabilistic method that enables a robot to acquire linguistic knowledge, including speech units, lexicons, grammar, and interpretation through human--robot embodied communication, in an unsupervised manner~\cite{Iwahashi2003}.
This integrated speech, visual, and behavioral information in a probabilistic framework.
This work was updated by Iwahashi et al.~\cite{Iwahashi2008}. The learning process was carried out online, incrementally, actively, and in an unsupervised manner.
On the basis of this work, Iwahashi et al. developed an integrated online machine-learning system called LCore, which combined speech, visual, and tactile information obtained through interactions, and enabled robots to learn beliefs regarding speech units, words, the concepts of objects, motions, grammar, and pragmatic and communicative capabilities~\cite{Iwahashi2010}.
These pioneering studies clearly demonstrated the possibility of the SER approach.

\subsection{Nonparametric Bayesian word segmentation}
In word discovery and segmentation tasks, the efficient management of a word inventory through the learning process is a fundamental computational problem. Although a robot can only memorize a finite number of words, there are potentially an infinite number of words in our society, i.e., in the human symbol system. The selection of an appropriate set of words constitutes a type of model selection problem, and usually requires a very large computational cost.
Recently, Bayesian nonparametrics have provided a sophisticated theoretical solution to this problem~\cite{HDP,HDP-HMM,Tenenbaum2011}. A nonparametric Bayesian language model, e.g., a hierarchical Pitman-Yor process language model (HPYLM), can assign an adequate probability to an infinite number of possible words using a fully Bayesian framework~\cite{teh2006b}.
On the basis of this framework, word segmentation methods that assume that there is an infinite number of possible words can be developed.
Goldwater proposed an HDP-based word segmentation method~\cite{Goldwater2008,Goldwater2009}.
Mochihashi et al. proposed a nested Pitman-Yor language model (NPYLM), in which a letter n-gram model based on a hierarchical Pitman-Yor language model is embedded in the word n-gram model~\cite{Mochihashi2009}. An efficient blocked Gibbs sampler, employing the forward filtering backward sampling procedure, was also introduced in that study. These methods have made it possible to discover words from transcribed phoneme sequences or text data without any recognition errors.

However, in practice phoneme recognition errors are inevitable, especially during the language acquisition phase.
In order to overcome this problem, several extensions have been proposed.
Neubig et al. extended the word segmentation methods of Mochihashi et al., making it possible to analyze phoneme lattices that are a type of expression of noisy speech recognition results~\cite{Neubig2012}. Heymann et al. modified the algorithm of Neubig et al., and proposed a suboptimal algorithm~\cite{Heymann2013,Heymann2014}.
Elsner et al. developed a learning method that jointly performs word segmentation and learns an explicit model of phonetic variation~\cite{Elsner2013}.
Hinoshita et al. solved a similar problem using MTRNN ~\cite{Hinoshita2011}.

Recently, several advanced machine-learning methods have attempted to learn words from acoustic data without a preexisting language model and acoustic model~\cite{Kamper2015,Brandl2008,Lee2015,NPB-DAA}. However, the problem remains a challenging topic.

\subsection{Integrative word discovery by robots}
It is difficult to discover words from uttered sentences using only distributed cues.
Researchers have developed robotic systems and computational models in which co-occurrence cues help to discover words using distributional cues.
Taguchi et al. proposed a method for the unsupervised learning of place-names, from information pairs that consist of spoken utterances and a mobile robot's estimated current location, without any prior linguistic knowledge other than a phoneme acoustic model~\cite{Taguchi2011}. They optimized a word list using a model selection method based on a description length criterion.
Araki et al. proposed an integrative computational model that involved MLDA and NPYLM~\cite{Araki2012}. Through MLDA, a robot can detect co-occurrence cues in the environment, and use the information to increase word segmentation performance.
It was shown that the iterative learning process comprising MLDA and NPYLM increased the word segmentation accuracy. However, they reported that the accuracy decreases as the phoneme recognition error rate increases~\cite{Araki2012}.
This implies that phoneme recognition errors and word segmentation errors should be mitigated simultaneously.
Nakamura et al. developed an integrated statistical model for word segmentation, speech recognition, and multimodal categorization, in order to overcome this problem. The robot in the experiment simultaneously formed object categories and learned related words from continuous speech signals and continuous visual, auditory, and haptic information, i.e., sensory--motor information, through an iterative learning process~\cite{Nakamura2014}.

Various word discovery methods which enable robots to obtain words and relationships between words and objects. However, situations in which robots can currently discover words are still limited. To simulate the robust word discovery process by human children in real-world environment, further studies will be required.

\section{Double Articulation Analysis}\label{sec6}
\subsection{Double articulation structure}
Double articulation is an important property of human language systems.
Chandler described double articulation as follows in a textbook on semiotics~\cite{Chandler2002}:
\begin{quote}
One of the most powerful ``design features'' of language is called double articulation (or ``duality of patterning''). Double articulation enables a semiotic code to form an infinite number of meaningful combinations using a small number of low-level units which by themselves are meaningless (eg., phonemes in speech or graphemes in writing). The infinite use of finite elements is a feature that about media, in general, has been referred to as ``semiotic economy.''
\end{quote}

Our speech signals and some semiotic time-series data are considered to have a double articulation structure. This means that a sentence can be decomposed into words, and a word can be decomposed into letters or phonemes.
Automatic speech recognition systems usually presume double articulation in speech signals.
A continuous speech signal is first segmented into phonemes, such as ``a,'' ``e,'' ``i,'' and ``s.'' Then, the phonemes are chunked into words, such as ``can,'' ``dog,'' and ``pen.'' 
A phoneme cannot usually act as a sign for an object in the sense of Peircean semiosis, but a word constitutes a sign, i.e., it has certain meaning.
Usually, the relationship between speech signals and phonemes is stored in a phonetic model and/or acoustic model, and the relationship between phonemes and words is stored in a language model. Therefore, direct word discovery from speech signals can be regarded as the analyzing of a latent double articulation structure embedded in speech signals in an unsupervised manner.

In addition to speech signals, other time-series data generated by humans may have a double articulation structure. If such time-series data exists, then the analysis of such data would contribute to our understanding of emergent symbol systems.
For this purpose, several researchers have developed a computational model that can automatically analyze double articulation structures~\cite{Taniguchi2011,Takano,takenaka12iros,tani_nolfi,NPB-DAA,Lee2015}.

\subsection{Segmentation of human bodily motion}
Human bodily motion is a candidate for doubly articulated time series data. Inamura et al. proposed the use of a left-to-right hidden Markov model (HMM) to recognize and reconstruct human bodily motion~\cite{Inamura2004}. Essentially, left-to-right HMMs are often used to model words in automatic speech recognition systems.
These computational models implicitly bridge speech recognition and human motion modeling.

There have been many previous studies on motion segmentation. However, the definition of a unit of motion has been unclear in many studies for a long time. Roughly speaking, some researchers have focused on physically elemental segments~\cite{rubin,Fod,kawashima,barbic,Li}, and others on semantically elemental segments~\cite{okada,kadone,Chippa}.
For example, when we semantically segment a baseball player's motion, ``pitching'' definitely becomes a candidate for a unit motion. However, pitching consists of several low-level segmental motions from the viewpoint of physical dynamics. If we segment the pitching motion according to the criterion that an elemental motion has linear dynamics, then the pitching motion will be segmented into several elemental motions, e.g., ``raising a knee'' and ``swinging an arm.'' However, these elemental motions seem to be meaningless, and have no special names. The two-layer hierarchical structure is similar to that existing in speech signals. We call a short-term motion that corresponds to phoneme a {\it segment}, and a long-term motion that corresponds to word a {\it chunk}. A chunk corresponds to a sequence of segments.

Takano et al. developed a large-scale database of human whole-body motion, and modeled the motion data using a large number of HMMs~\cite{Takano,Takano2010}. They roughly clustered the given motions, and constructed many left-to-right HMMs corresponding to meaningful motions. They hierarchically clustered the HMMs again, and obtained a large motion database. An online incremental learning method was also provided by Kuli\'{c}~\cite{Kulic2012}.
They implicitly assumed double articulation in human bodily motion.
Based on the database, they developed a machine translation system that can translate continuous human motion into a sentence, and vice versa~\cite{Takano2008a,Takano2009}.

Taniguchi et al. proposed a double articulation analyzer (DAA) by combining a sticky hierarchical Dirichlet process-HMM (sticky HDP-HMM) and NPYLM~\cite{Fox2009,Taniguchi2011}. The DAA explicitly assumes double articulation, and infers the latent letters, i.e., the segment or phoneme, and the latent words, i.e., the words or segments, in an unsupervised manner. These two nonparametric Bayesian models, sticky HDP-HMM and NPYLM, were sequentially applied to the target data, and a two-layered hierarchical structure was inferred in the DAA.
They applied the DAA to human motion data, to extract unit motions from unsegmented human motion data.
\begin{figure}[t]
  \centering
  \includegraphics[width = 1.0\linewidth]{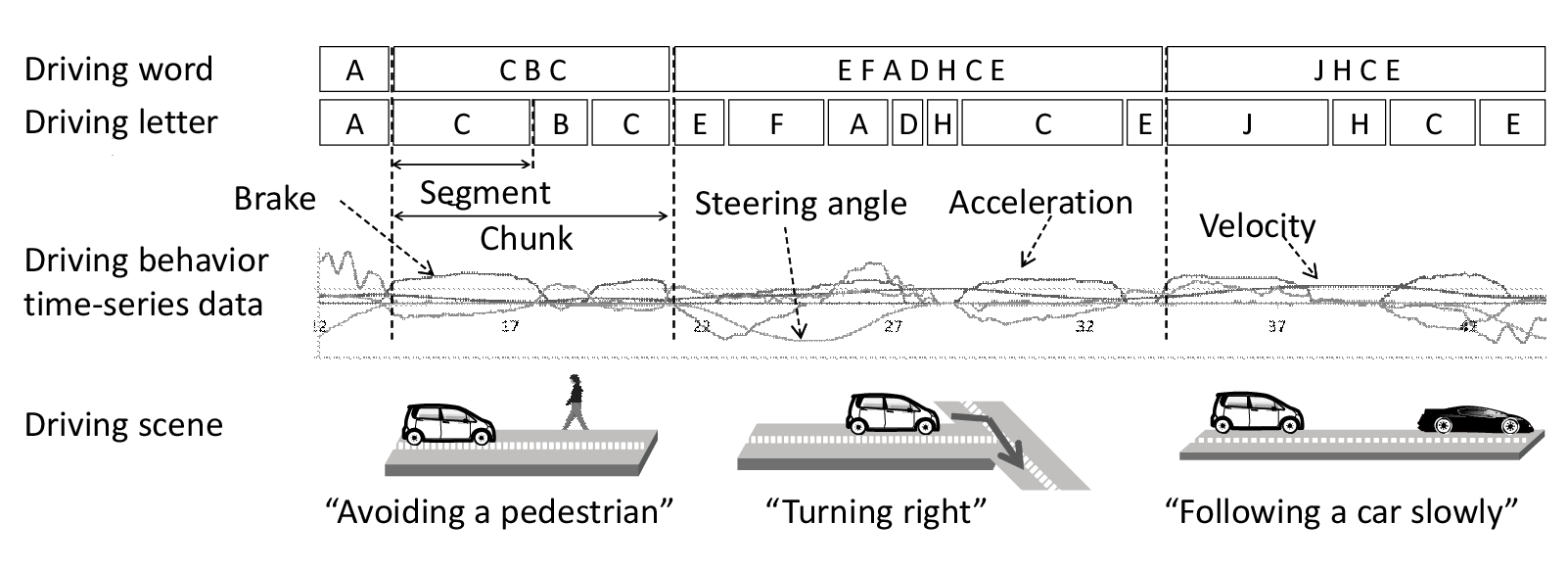}
  \caption{Double articulation in driving behavior.}
  \label{fig:da_in_driving}
\end{figure}

\subsection{Modeling driving behavior data}
Data on driving behavior represents another candidate for doubly articulated time-series data.
A meaningful chunk of driver behavior seems to consist of sequences of simple segments. Figure~\ref{fig:da_in_driving} presents an example. In this figure, when a driver ``turns right at an intersection,'' he/she ``steps on the brake,'' ``turns the steering wheel to the right,'' ``steps on the accelerator,'' ``turns the steering wheel back to center,'' and ``steps on the accelerator'' again as a sequence of physically elemental driving behaviors.

The DAA has been utilized for various application, such as segmentation~\cite{takenaka12iros}, prediction~\cite{taniguchi12iv,IEEE-ITS}, data mining~\cite{genki12sii}, topic modeling~\cite{bando13iv,bando13iros}, and video summarization~\cite{takenaka12mm}.
Through experiments in a series of studies, it has become clear that driving behavior data has a double articulation structure. For example, a prediction method based on the DAA has outperformed conventional methods in a diving behavior prediction task~\cite{IEEE-SMC}. This implies that the assumption of double articulation is appropriate. Taniguchi et al. called the latent states corresponding to semantically elemental driving behaviors {\it driving words}, and those corresponding to physically elemental driving behaviors {\it driving letters}.
Recently, the DAA was applied to large-scale driving behavior data, and its effectiveness for driving behavior analysis has been verified~\cite{Mori2015}.

\subsection{Predicting long-term sensory--motor information}
A third candidate for doubly articulated time-series data is given by sensory--motor information flow. Unlike speech signals, sensory--motor information flow data comprises sensor information as well as motor information. Modeling sensory--motor time-series data for a robot means estimating the forward dynamics of the system that the robot confronts. Many modular learning architectures, e.g., MOSAIC (modular selection and identification for control), HAMMER (hierarchical, attentive, multiple models for execution and recognition), and the dual schemata model, have been proposed to model switching forward dynamics~\cite{Wolpert1998,Demiris2006,Taniguchi2003,Taniguchi2004,Taniguchi2008}. Most of these involve forward-inverse models in the learning system. Methods for imitation learning, reinforcement learning, and the emergence of communication have been proposed on the basis of such modular learning architectures~\cite{samejima,social_mosaic,Doya2002,Taniguchi2007}. 

In dynamic environments, the forward dynamics of a robot change intermittently. Such contextual information tends to have a certain structure. Tani et al. proposed a the use of hierarchical mixture of RNNs~\cite{tani_nolfi}. In their experiment, a longer context was coded into the activations of the context nodes of the RNNs at a higher level, and a shorter context was coded into those at a lower level.
In other words, their model captured the double articulation structure in the environmental dynamics.
Hierarchical MOSAIC and HAMMER architectures have also been proposed as computational models that could capture such hierarchical structures~\cite{haruno2003hmm,Demiris2006}. MTRNN and recurrent neural networks with parametric bias (RNNPB) are candidates to model contextually changing forward dynamics~\cite{Tani2004, Heinrich2014a, Tani2014, Murata2014}

\subsection{Direct word discovery from speech signals}
As mentioned above, the double articulation analysis is deeply related to direct word discovery from an acoustic speech signal. Several studies have recently been carried out in relation to this difficult problem.
The direct application of the DAA proposed by Taniguchi et al.~\cite{Taniguchi2011} is one possible approach; however, poor results are expected to be obtained.
In that approach, the DAA simply uses the two nonparametric Bayesian methods sequentially. These are not integrated into a single generative model. Therefore, if there are many recognition or categorization errors in the result of the first segmentation process by sticky HDP-HMM, then the performance of the subsequent unsupervised chunking by NPYLM deteriorates.

To overcome this problem, Taniguchi et al. proposed a nonparametric Bayesian double articulation analyzer (NPB-DAA), which is a two-layered generative model~\cite{NPB-DAA}. The generative model represents a complete data generation process that has a double articulation structure. An efficient blocked Gibbs sampler was also derived in the same study. They showed that the NPB-DAA could automatically find a word list from vowel speech signals directly and completely, in an unsupervised manner. Additional approaches have recently been proposed~\cite{Lee2015,Kamper2015}.

It is interesting that superficially different time-series data generated from human behavior can be analyzed using almost the same computational model. In addition, the characteristic of double articulation is satisfied. That is, the elements in the first layer, i.e., the phonemes and segments, are meaningless, and the elements in the second layer, i.e., words and chunks, are meaningful.
This suggests that different examples of such time-series data potentially share the same computational processes in our brain. In addition, we hypothesize that these are profoundly related to the nature of our emergent symbol system.

\section{Further Topics}\label{sec7}
In order to construct a computational model that describes an entire emergent symbol system, and to develop a robot that can communicate and collaborate with humans in a long-term manner, there are many other challenges in the field of SER to consider and overcome. In this section, we describe some of them.
\subsection{Mutual belief system}
An utterance is not usually interpreted ``as it is'' by another person to whom it is told.
A mutual belief system always affects a person's interpretation. Roy provides a coffee scenario as an example in his survey paper~\cite{Roy}.
Imagine a situation in which a cup of coffee is served to a customer by a waitperson, and the customer says, ``This coffee is cold.''
In this case, the referential meaning of the utterance is the fact that the temperature of the coffee is low in the sense of thermodynamics, but the functional meaning is ``please get me a hotter coffee.''
The speech act conveys a meaning interpreted by referring to the physical situation shared by the communication partners. This means that the mutual belief system is important in generating natural sentences and interpreting uttered sentences.
Roy emphasized this aspect of language for solving the symbol grounding problem.

A pioneering study that develops a constructive model involving a mutual belief system was presented by Iwahashi~\cite{Iwahashi2003}. Iwahashi introduced a belief function that represents a mutual belief of a robot. The belief function contains several belief modules, i.e., speech, object images, motions, motion--object relationships, and behavioral contexts. The various external and internal contexts are taken into consideration to infer the speaker's intention. The robotic system is truthfully an embodied natural language processing system that can take various contexts that an ordinal amodal natural processing system cannot access. 
Zuo et al. applied this model for detecting robot-directed speech~\cite{Zuo2010}. Sugiura et al. proposed a method for estimating the ambiguity in commands by introducing an active learning scheme to the conversation system, based on the mutual belief~\cite{Sugiura2011}.

A mutual belief is part of an emergent symbol system, and applies ``constraints'' not only to the interpretation of utterances, but also to the generation of our speech.
Context is a crucial element of our natural dialog but is rarely taken into consideration in natural language processing. The SER approach is promising in this topic as well.

\subsection{Active perception and learning}
A robot that has the potential to be an element of an emergent symbol system, i.e., a member of our semiotic society, must be able to explore its environment, acquire knowledge, and communicate with people autonomously.
Active perception and active learning are two of the most important capabilities of humans for achieving life-long development and communication~\cite{Denzler2002,Krainin2011a,Eidenberger2010a,VanHoof2012a,Rebguns2011,Fox,Roy,Stachniss,Gouko,Saegusa2011,Ji2006,Tuci2010,Natale,Schneider2009a}.

Denzler et al. proposed an information theoretic action selection method to gather information that conveys the true state of a system through an active camera~\cite{Denzler2002}.
They used mutual information (MI) as a criterion for action selection.
Krainin et al. developed an active perception method that made a mobile robot manipulate objects to build three-dimensional surface models of the objects~\cite{Krainin2011a}.
Their method determines when and how a robot should grasp an object on the basis of the information gain (IG) criterion.

Modeling and recognizing a target object, as well as modeling a scene and segmenting objects from that scene, are important abilities for a robot in a realistic environment.
An active perception planning method for scene modeling in a realistic environment was proposed by Eidenberger et al.~\cite{Eidenberger2010a}. A partially observable Markov decision process (POMDP) formulation was used to model the planning problem, and the differential entropy was introduced as part of the reward function.
Hoof et al. proposed an active scene exploration method. An autonomous robot is able to segment a scene into its constituent objects by actively interacting with the objects using this method~\cite{VanHoof2012a}. They used IG as a criterion for action selection.
InfoMax control for acoustic exploration was proposed by Rebguns et al.~\cite{Rebguns2011}.
In general, IG is mostly applied in active perception and learning~\cite{Denzler2002,Krainin2011a}. Taniguchi et al. developed an optimal active perception scheme for multimodal categorization using MLDA on the basis of the IG criterion.
Localization, mapping, and navigation are also important targets of active perception~\cite{Fox,Roy,Stachniss}. In addition, various other studies on active perception have been conducted~\cite{Gouko,Saegusa2011,Ji2006,Tuci2010,Natale,Schneider2009a}.

In the context of reinforcement learning (RL), intrinsic motivation has been studied~\cite{Singh2005}. In reinforcement learning, an agent autonomously learns its policy, i.e., controller, to maximize the expected cumulative rewards. In related studies, internal reward systems are taken into consideration as well as external reward systems. Schmidhuber presented a survey on RL studies considering intrinsic motivation, and proposed a formal theory of fun, intrinsic motivation, and creativity~\cite{Schmidhuber2010}.  

When we develop an autonomous robot, the design of the intrinsic motivation and explanatory behavior, active perception, and in particular the IG criterion provide cues for the problem. The effectiveness of the IG criterion tells us that ``curiosity'' is properly treated computationally, in contrast with other emotions.

Sugiura et al. used active perception to determine the utterances of a robot~\cite{Sugiura2011}. When we talk to each other, we anticipate a reply, i.e., some information, from the other person. Therefore, we might generate a sentence so as to maximize the information gain or expected reward. Communication always involves some kind of decision-making problems. 
Active perception and learning will become more and more important in the wide range of decision-making problems concerning robots.

\subsection{Compositionality and semantics}
The hierarchical structure placed on segmented words that are extracted on the basis of double articulation analysis enables us to generate various meaningful sentences. A syntax is a rule that produces a meaningful sequence of words. From the viewpoint of an emergent symbol system, an important problem can be phrased as ``how can combined words have adequate meanings for an embodied and situated agent?''  
An important research topic concerns computational models for grounding semantic composition~\cite{Daoutis2014}.
Classically, the principle of compositionality, which is also called Frege's principle, has been widely recognized. This is a principle stating that the meaning of a complex expression is determined by the meanings of its constituent expressions and the syntactic rules used for combining them~\cite{Morris1994}. 
A bottom-up approach to the principle of compositionality also represents an important topic in symbol emergence in robotics.

The notion of combinatoriality has been studied in a constructive manner. Sugita et al. and Ogata et al. developed a robotic system and neural network architecture that can simultaneously learn sentences and behaviors~\cite{Sugita2005,Tani2004,Ogata2007}. For this, they used an RNNPB. It was shown that compositionality emerges on the network in a distributed manner. Tuci et al. also introduced neurodynamic models that deal with compositionality problems in language and behavior association learning, and the learning of goal-directed actions~\cite{Tuci2011,Sandamirskaya2013}. 
Hinaut et al. applied reservoir computing, which is a kind of RNN, to make a robot acquire and produce grammatical constructions~\cite{Hinaut2014,Hinaut2015}.
Tani and Cangelosi et al. have presented a comprehensive survey of related studies~\cite{Tani2014,Cangelosi2010}.

Recently, distributional models of semantics, including word2vec, have been given attention~\cite{Mikolov2013,Mikolov2013a,Mikolov2013b}. For example, Mikolov showed that the relationship between a country and a capital city can automatically extracted from an unlabeled text dataset only by a training predictor, on the basis of a skip-gram and recurrent neural net language model (RNNLM).
Le et al. proposed an unsupervised machine-learning method, called paragraph vector, that can estimate fixed-length feature representations from variable-length sections of texts, e.g., sentences, paragraphs, and documents~\cite{Le2014}. 
Many preliminary studies exist concerning compositionality and semantics. Incorporating both embodied cognition and formal language structure must be important to construct a robot that can understand uttered sentences in the real word.  @
The connection of such learning methods to a series of studies in SER also represents an important topic.

\section{Conclusion}\label{sec8}
In this paper, we have provided an introduction to emergent symbol systems and surveyed the research field of SER. Semioticians sometimes call humans ``Homosignificans,'' which means meaning-makers~\cite{Chandler2002}. Comprehending signs from natural or artificial environments and applying semiosis in the mind are human characteristics. When developing an autonomous robot that can engage in long-term communication and collaboration with people, the robot must adapt to the human symbol system.
To provide a philosophical framework for the diversity and dynamics of a symbol system, we introduce a concept---the emergent symbol system---that constitutes a basic assumption in SER. People can acquire language through physical interactions with their environment and semiotic communication with other people (see Figure~\ref{fig:ses}).
This phenomenon is comprehended as a type of assimilation process, in which a personal symbol system that is supported by an internal representation system becomes coupled with the emergent symbol system. To achieve such assimilation, the person must have the capability to learn the language in an unsupervised manner. For robots, this must be same.
To achieve long-term interaction with people, a robot has to have the capability to learn the language in an unsupervised manner, so that the robot's symbol system becomes coupled with the emergent symbol system that the target society has.
Therefore, it is crucially important to computationally understand how humans can learn a symbol system and obtain semiotic skills through their autonomous mental development.

Many challenges have been investigated in relation to the construction of robotic systems and machine-learning methods that can obtain some parts of language through the embodied multimodal interaction with the environment. In order to understand human social interactions, and develop a robot that can smoothly communicate with human users, it is fundamentally important to understand symbol systems that change dynamically on the basis of the embodied cognition of participants in a constructive manner.

In this paper, we introduced the research field called SER. This represents a constructive approach towards an emergent symbol system. The emergent symbol system is socially self-organized through both semiotic and physical interactions with autonomous cognitive developmental agents, i.e., humans and developmental robots.
Among the numerous fields connected with SER, we have described some specific topics in this paper, such as multimodal categorization, word discovery, and double articulation analysis.
SER presents various future challenges involving acquiring lexicons, learning syntax, obtaining skills using metaphors, learning pragmatics, and being able to generate appropriate sentences given a particular context.

The majority of studies relating natural language processing and linguistics have only treated documents. However, we have to communicate and collaborate with other agents, including people and robots, in a real-world environment. The appropriateness of emergent symbol systems and robotic systems must be evaluated in relation to embodied cognition, context, and collaboration.
Real-world collaborative tasks should be considered in order to evaluate them. In this context, some competitions including real-world human--robot interaction must be effective  for facilitating researchers to study embodied cognitive systems and emergent symbol systems, and for evaluate appropriateness of the developed systems. RoboCup@Home is an obvious candidate~\cite{Stuckler2009,Iocchi2015,Nakamura2012b,Stuckler2012}. 

However, the learning of all of the knowledge required for communication and collaboration through situated and embodied interactions requires very large costs and time. For further research, some pseudo-real-world environment will be required to increase the speed of our research. For example, Inamura et al. have developed the SIGVerse, a SocioIntelliGenesis simulator that enables human--robot interactions in a virtual world~\cite{Inamura2010}. Cloud-based semiotic and physical interactions will also be important components of further studies in SER.

SER is still an emerging research field, but one that shows promise. Further research on SER will push long-term human--robot interaction forward, and provide a new understanding of human intelligence.

\section*{Acknowledgements}
This research was partially supported by a Grant-in-Aid for Young Scientists (B) 2012-2014 (24700233) and a Grant-in-Aid for Young Scientists (A) 2015-2019 (15H05319) funded by the Ministry of Education,
Culture, Sports, Science, and Technology, Japan. This research was supported by CREST, JST.

\bibliographystyle{tADR}
\bibliography{ar}
\label{lastpage}

\end{document}